\documentclass{article}

 \PassOptionsToPackage{numbers, compress}{natbib}

\usepackage[preprint]{nips_2018}

\usepackage[utf8]{inputenc} 
\usepackage[T1]{fontenc}    
\usepackage{hyperref}       
\usepackage{url}            
\usepackage{booktabs}       
\usepackage{amsfonts}       
\usepackage{nicefrac}       
\usepackage{microtype}      
\usepackage{graphicx}
\usepackage{amsmath}

\title{SpaceNet: A Remote Sensing Dataset and Challenge Series}

\author{
  Adam Van Etten \\
  In-Q-Tel CosmiQ Works\\
  Arlington, VA 22201 \\
  \texttt{avanettten@iqt.org} \\
  \And  
  Dave Lindenbaum \\
  In-Q-Tel CosmiQ Works \\
  Arlington, VA 22201 \\
  \texttt{dlindenbaum@iqt.org} \\
  \And 
  Todd Bacastow \\
  Radiant Solutions \\
  Herndon, VA 20171 \\
  \texttt{todd.bacastow@radiantsolutions.org}
}

\begin{document}

\maketitle

\begin{abstract}
Foundational mapping remains a challenge in many parts of the world, particularly in dynamic scenarios such as natural disasters when timely updates are critical.  Updating maps is currently a highly manual process requiring a large number of human labelers to either create features or rigorously validate automated outputs.  We propose that the frequent revisits of earth imaging satellite constellations may accelerate existing efforts to quickly update foundational maps when combined with advanced machine learning techniques.  Accordingly, the SpaceNet partners (CosmiQ Works, Radiant Solutions, and NVIDIA), released a large corpus of labeled satellite imagery on Amazon Web Services (AWS) called SpaceNet.  The SpaceNet partners also launched a series of public prize competitions to encourage improvement of remote sensing machine learning algorithms.  The first two of these competitions focused on automated building footprint extraction, and the most recent challenge focused on road network extraction.  In this paper we discuss the SpaceNet imagery, labels, evaluation metrics, prize challenge results to date, and future plans for the SpaceNet challenge series.  
\end{abstract}

\section{Background}

The commercialization of the geospatial industry has led to an explosive amount of data being collected to characterize our changing planet. One area for innovation is the application of computer vision and deep learning to extract information from satellite imagery at scale. To this end, CosmiQ Works, Radiant Solutions, and NVIDIA partnered to release SpaceNet data as a Public Dataset on AWS \cite{sn_s3}.

Today, map features such as roads, building footprints, and points of interest are primarily created through manual mapping techniques. We believe that advancing automated feature extraction techniques will serve important downstream uses of map data, such as humanitarian and disaster response. Furthermore, we believe that solving this challenge is an important stepping stone to unleashing the power of advanced computer vision algorithms applied to a variety of remote sensing data applications in both the public and private sectors.

The first two SpaceNet challenges focused on building footprint extraction from satellite imagery.  The third challenge addressed another foundational geospatial intelligence problem, road network extraction.  The ability to create a road network is an important map 
feature (particularly if one is able to use this road network for routing purposes), while building footprint extraction serves as a useful proxy for population density \cite{zhangpop}, \cite{popnature}, \cite{popplos1}. The potential application of automated building footprint and road network extraction ranges from the determination of optimal aid station locations during an epidemic in under mapped locations, to the establishment of an effective logistical schema in a disaster stricken region.    There is a demonstrated need for both automated building footprint extraction as well as automated road network extraction, as the Humanitarian OpenStreetMap Team's Tasking Manager \cite{hotosm}
currently has scores of open tasks for both roads and buildings. 

To address these challenges (and inspired by the ImageNet \cite{imagenet} model), the SpaceNet team released a large corpus of labeled satellite imagery in conjunction with public challenges aimed to increase the utility of satellite imagery.  
All SpaceNet data is distributed under a Creative Commons Attribution-ShareAlike 4.0 International License and is hosted as a public dataset on AWS and can be downloaded for free.
 The future vision for SpaceNet is the continued expansion of automated capabilities available to foundational mapping
practitioners and users.  This will be accomplished by releasing high quality labeled datasets, and running targeted public challenges to encourage the development of algorithms designed to solve increasingly complex geospatial problems. Two future challenges are already in the execution stage, and will be discussed further in Section \ref{conclusions}.


\section{Source Imagery and Labels}

\subsection{Existing Datasets}

Existing publicly available labeled overhead or satellite imagery datasets tend to be relatively small, or labeled with lower fidelity than desired for foundational mapping.  For example, the ISPRS semantic labeling benchmark \cite{isprs_sem}
dataset contains high quality 2D semantic labels over two cities in Germany and covers a compact area of 4.8 km$^2$; 
imagery is obtained via an aerial platform and is 3 or 4 channel and  5-10cm in resolution.  The TorontoCity Dataset \cite{torontocity} contains high resolution 5-10cm aerial 4-channel imagery, and $\sim700$ km$^2$ of coverage; building and roads are labeled at high fidelity (among other items), but the data has yet to be publicly released.  The Massachusetts Roads Dataset \cite{MnihThesis} contains 3-channel imagery at 1 meter resolution, and $2600$ km$^2$ of coverage; the imagery and labels are publicly available, though labels are scraped from OpenStreetMap and not independently collected or validated. Another useful overhead dataset (though not so relevant to foundational mapping) is the COWC dataset \cite{cowc} of cars, with 15cm aerial imagery collected over six different geographic regions; labels consist of a point at the centroid of each car. A recently released satellite imagery dataset is xVIEW \cite{xview_data}, which contains 1400 km$^2$ of 30 cm satellite imagery; labels, however, consist of bounding boxes that are not ideal for foundational mapping of buildings or roads.  

\subsection{Other Satellite Imagery Competitions}
	A number of public challenges using satellite imagery have been run in the last couple of years.  While SpaceNet pre-dates and helped inspire some of these challenges, the results of these challenges are of great benefit to the community and very informative to future SpaceNet competitions.  
	The Dstl Satellite Imagery Feature Detection \cite{dstl} ran on Kaggle shortly after the completion of SpaceNet 1, and sought to segment multiple classes in high resolution satellite imagery.  
	The IARPA Functional Map of the World \cite{fmow} challenge and dataset aimed to classify region proposals based on building type. 
	The DIUx xView Detection Challenge \cite{xview} subsequently released a large dataset of 60 bounding box object classes and approximately 1 million instances.  
	In a similar vein to SpaceNet, the DeepGlobe Satellite Challenge at CVPR 2018 \cite{deepglobe} consisted of three distinct challenges: road segmentation, building detection, and land cover classification.  The SpaceNet team assisted with this challenge, and much of the DeepGlobe data stemmed from the SpaceNet repository.  Finally, the ISPRS Semantic Labeling Contest  \cite{isprssemlab} pre-dates all of these and provided an important benchmark for computer vision performance on remote sensing data.

\subsection{Challenge 1 - Rio De Janeiro Building Footprints}
The first SpaceNet challenge in 2016 aimed to extract building footprints from the DigitalGlobe WorldView 2 satellite imagery at 50cm resolution.  Imagery consists of 8-band multispectral imagery at 1m resolution, as well as pan-sharpened red-green-blue (RGB) imagery at 50cm resolution.  Released imagery was created from a mosaic of multiple images, and covers 2544 square kilometers.  The challenge imagery 
was split into 200 meter tiles using the SpaceNet utilities python package \cite{sn_utils},
with 60\% of the data released for training, 20\% for testing, and 20\% reserved for validation. 

\begin{figure}
  \centering
     \includegraphics[width=0.9\linewidth]{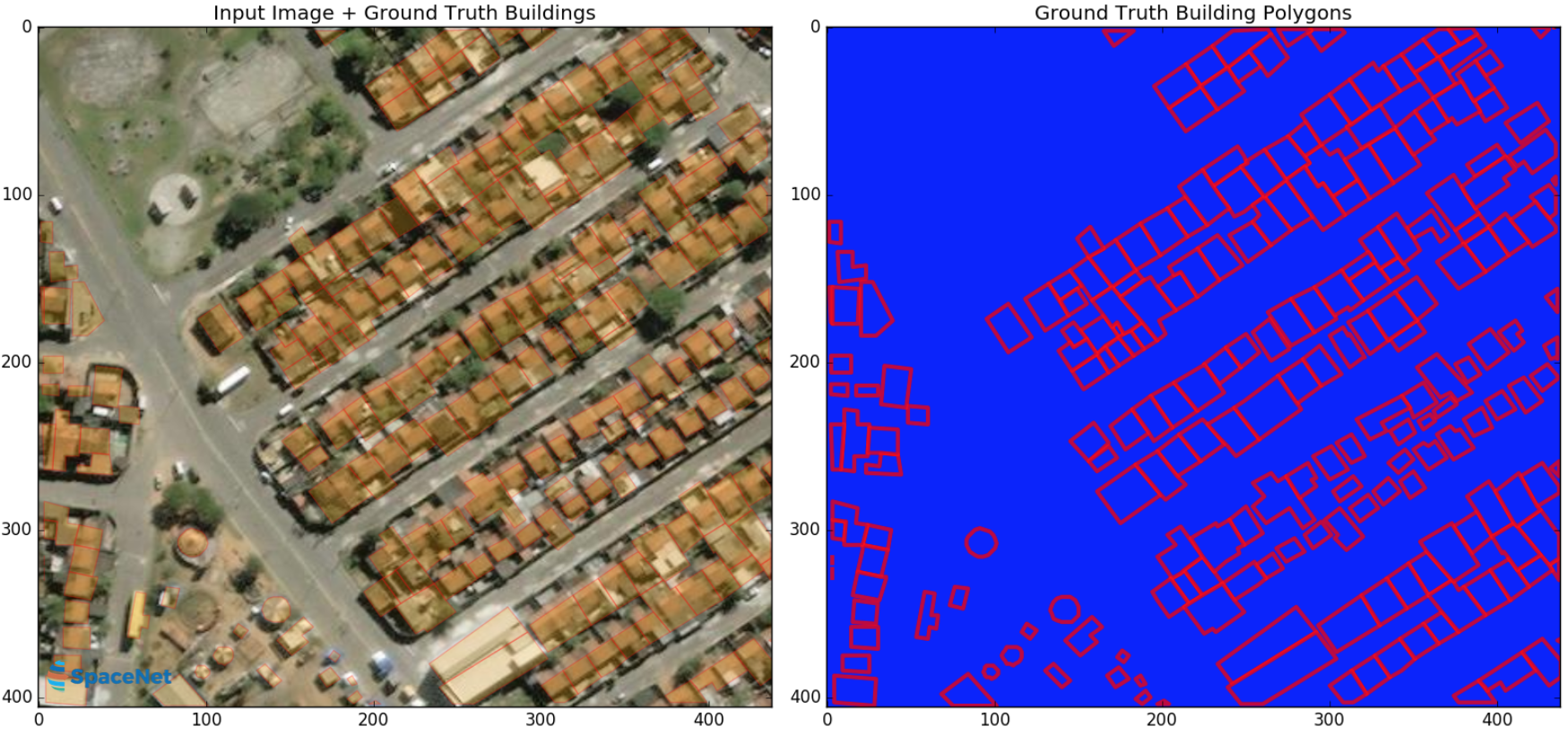}
  \caption{200m SpaceNet chip over Rio de Janeiro and attendant building labels}
  \label{fig:sn_data0}
\end{figure}

For Challenge 1 in the Rio de Janeiro Area of Interest (AOI) over 300,000 building footprints were labeled.
A GIS team at DigitalGlobe (now Radiant Solutions) provided a polygon footprint for each building, which were initially extracted through semi-automated means and subsequently improved by hand.  
Any partially visible rooftops were approximated to represent the shape of the building. Adjoining buildings were marked individually as unique structures (i.e.~each street address was marked as a unique building), see Figure \ref{fig:sn_data0}.

\begin{table}
  \caption{SpaceNet imagery and building label details}
  \label{tab:data_area}
  \centering
   \begin{tabular}{llllll}
    \toprule
AOI 	& Area  & \# Buildings & GSD & Sensor & Date \\
	& (km$^2$) & (Polygons) & (cm) & & \\
    \toprule
Rio & 2,544 &	382,534	& 50 &	WorldView-2	& 2011--2014 \\
Las Vegas & 216 & 151,367 &	30	 & WorldView-3 & 2015-10-22  \\
Paris & 1,030 & 23,816 & 30 & WorldView-3 & 2016-02-29  \\
Shanghai & 1,000 & 92,015 & 30 & WorldView-3 & 2015-06-06  \\
Khartoum & 765 & 35,503 & 30 & WorldView-3 & 2015-04-13  \\

    \bottomrule
  \end{tabular}
\end{table}

\subsection{Challenge 2 - Las Vegas, Paris, Shanghai, Khartoum Building Footprints}

The second SpaceNet challenge aimed to extract building footprints from the DigitalGlobe WorldView 3  satellite in a continuous image strip. The source imagery is distributed as a Level 2A standard product that has been radiometrically and sensor corrected, and normalized to topographic relief using a coarse digital elevation model (DEM). It contains the original panchromatic band, the 1.24m resolution 8-band multi-spectral 11-bit geotiff, and a 30cm resolution Pan-Sharpened 3-band and 8-band 16-bit geotiff. Table \ref{tab:data_area} provides collection details for each SpaceNet Area of Interest (AOI). 

The dataset includes four areas: Las Vegas, Paris, Shanghai, and Khartoum. The labeled dataset consists of 24,586 scenes of size $200 \,\rm{m} \times 200 \,\rm{m} \, (650 \,\rm{px} \times 650 \,\rm{px})$ 
containing 302,701 building footprints across all areas, and are both urban and suburban in nature. 
The dataset was split 60\%/20\%/20\% for train/test/validation.
Each area is covered by a single image strip, which ensures that sun, satellite, and atmospheric conditions are consistent across the entire scene.
 A GIS team at DigitalGlobe (now Radiant Solutions) fully annotated each scene to within 5 pixels for building polygon corners (see Appendix A). The labels went through a rigorous QA/QC process, with one expert labeling a specific area and then a second expert performing validation on $100\%$ of the areas. For buildings the final product was then inspected against topology errors to ensure that building footprints were closed and polygons did not overlap. With human-based annotation some small errors are inevitable, especially for rural areas, and we leave the analysis of annotator disagreement for future work.



\subsection{Challenge 3 - Las Vegas, Paris, Shanghai, Khartoum Road Extraction}

Challenge 3 used the imagery from Challenge 2,  only tiled into 400m chips. 
As with the previous challenges the chipped dataset was randomly split 60\%/20\%/20\% for train/test/validation.

\begin{figure}
  \centering
     \includegraphics[width=0.9\linewidth]{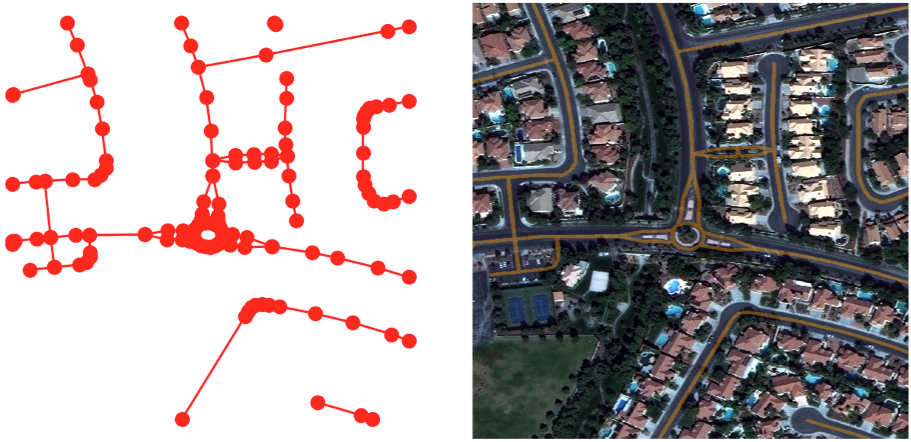}
  \caption{400m SpaceNet chip of Las Vegas.  Left: SpaceNet GeoJSON road label. Right: RGB image overlaid with road centerlines (orange).}
  \label{fig:sn_data1}
\end{figure}

CosmiQ Works worked with Radiant Solutions to label the road networks for the entire area of the existing SpaceNet Round 2 Building Dataset (see Figure \ref{fig:sn_data1}). Because the competition was designed to enable the creation of routable road networks, a labeling schema based on OpenStreetMap (OSM) guidelines was established to ensure ground truth that would be usable by open source routing tools. In addition to the digitizing of road- centerlines, four other attributes were recorded: 1) road type, 2) surface type, 3) bridge and 4) lane number. 
	 A GIS team at DigitalGlobe (now Radiant Solutions) fully annotated each road centerline within 7 pixels (see Appendix B). The roads labels went through the same QA/QC process as SpaceNet Challenge 2
	 Finally each completed area of interest was inspected and validated, with all reported dangling roads or missed connections corrected. The complete label guidelines can be found in Appendix B. Table \ref{tab:road_labels} provides a summary of road labels by road type and area of interest.

\begin{table}
  \caption{SpaceNet road labels}
  \label{tab:road_labels}
  \centering
   \begin{tabular}{llllll}
    \toprule
Road Type & AOI 2 Vegas & AOI 3 Paris & AOI 4 Shanghai & AOI 5 Khartoum & Total \\
\toprule
Motorway & 115.0 km & 9.0 km & 102.1 km & 13.4 km & 239.5 km \\
Primary & 364.7 km & 14.3 km & 191.5 km & 98.4 km & 668.9 km \\
Secondary & 416.9 km & 58.1 km & 500.7 km & 65.8 km & 1041.5 km \\
Tertiary & 2.9 km & 10.5 km & 33.6 km & 67.9 km & 115.0 km \\
Residential & 1645.6 km & 232.4 km & 938.3 km & 484.5 km & 3301.3 km \\
Unclassified & 1138. km & 95.4 km & 1750.8 km & 164.5 km & 3148.7 km \\
\toprule
Total & 3685.0 km & 425.3 km & 3536.9 km & 1029.5 km & 8676.6 km \\
     
    \bottomrule
  \end{tabular}
\end{table}

\subsection{Additional Data}

In addition to the data detailed above, SpaceNet also hosts data from previous competitions such as the IARPA Functional Map of the World Competition \cite{fmow} and the Urban 3D Challenge Dataset \cite{Urban3D2017}.  These datasets diversify the SpaceNet data corpus, with the goal encouraging new avenues of machine learning research.

\section{The SpaceNet Buildings Metric}

The aim of the first two SpaceNet challenges was to extract building footprints from high-resolution satellite imagery, which is an object detection problem.  Accordingly, we adopt a similar approach to ImageNet \cite{imagenet} and adopt a metric based upon the scale invariant intersection over union (IoU) metric, also known as the Jaccard index, which provides a measure of the overlap between two objects by dividing the area of the intersection by the area of the union. 

\begin{equation}
\label{eqn:iou}
IoU(A,B) = area(A \cap B) / area( A \cup B)
\end{equation}



As with Equation 5 of ImageNet, we select an IoU threshold of $\geq 0.5$ to denote a true positive detection.  
One additional feature that SpaceNet adopts from the ImageNet competition 
is the notion that each labeled region can have at most one true positive associated with that labeled region. This feature is implemented by a sequential search for a true positive sorted by decreasing IoU values. If a true positive is found, then the pair (the label and the proposed region)  are removed from the sequence and the search continues. 

With an object detection algorithm, the performance of an algorithm should depend on how many objects the algorithm detects (true positives)
how many objects it fails to detect (false negatives),
and how many non-objects it detects (false positives).
Accordingly, for each scene in the SpaceNet data corpus, we compute the total number of true positives, false positives (proposals with IoU $<$ 0.5), and false negatives (ground truth buildings without a valid proposal).  In a given city, we sum the true and false positives and negatives to compute total precision and recall scores; the city-wide F1 score is given by the harmonic mean of precision and recall:




\begin{equation}
\label{eqn:f1}
F1 = 2 \times (Precision \times Recall) / (Precision + Recall) 
\end{equation}

The total SpaceNet Buildings Metric is tabulated via the arithmetic mean of the F1 scores of each city.

\section{APLS: The SpaceNet Roads Metric}

The third SpaceNet competition aimed to extract road networks from satellite imagery.  Historically, pixel-based metrics have often been used to assess the quality of road proposals, though such metrics are suboptimal for a number of reasons that we discuss below.  Two graph-theoretic metrics have also been used (albeit with less frequency than pixel-based metrics), though these metrics still don't capture quite what was intended with SpaceNet 3.  Accordingly, we developed a novel metric (Average Path Length Similarity: APLS) \cite{apls} to measure similarity between ground truth and proposal road graphs. 

\subsection{Pixel-Based Metrics}

Road localization from overhead imagery has often been treated as an image segmentation problem (i.e.~identifying which pixels in an image belong to which class).  In segmentation problems, each prediction pixel is either a true positive,
false positive,
or false negative.
Three commonly used evaluation metrics are discussed below.

\begin{description}

\item[Pixel-Based IoU]

For object detection problems a threshold IoU value is often defined, above which the prediction is assumed to be a true positive.  For pixel-based approaches, the number of true positive pixels forms the intersecting area, and the sum of true positive, false positive, and false negative pixels forms the union.  IoU is subsequently calculated from Equation \ref{eqn:iou}.

\item[Pixel-Based F1 Score]
Each prediction pixel is either a true positive, false positive, true negative, or false negative. These can be combined into an F1 score via Equation \ref{eqn:f1}.  

\item[Relaxed F1]
This is computed by first calculating the relaxed precision and relaxed recall.  Relaxed precision is the fraction of predicted road pixels that are within a predetermined number of road pixels (Q) of a true road pixel \cite{relaxedf1}.  Relaxed recall is the fraction of true road pixels within Q pixels of a predicted road pixel.  The F1 score is subsequently tabulated via Equation  \ref{eqn:f1}.

\end{description}

\begin{figure}
  \centering
     \includegraphics[width=0.9\linewidth]{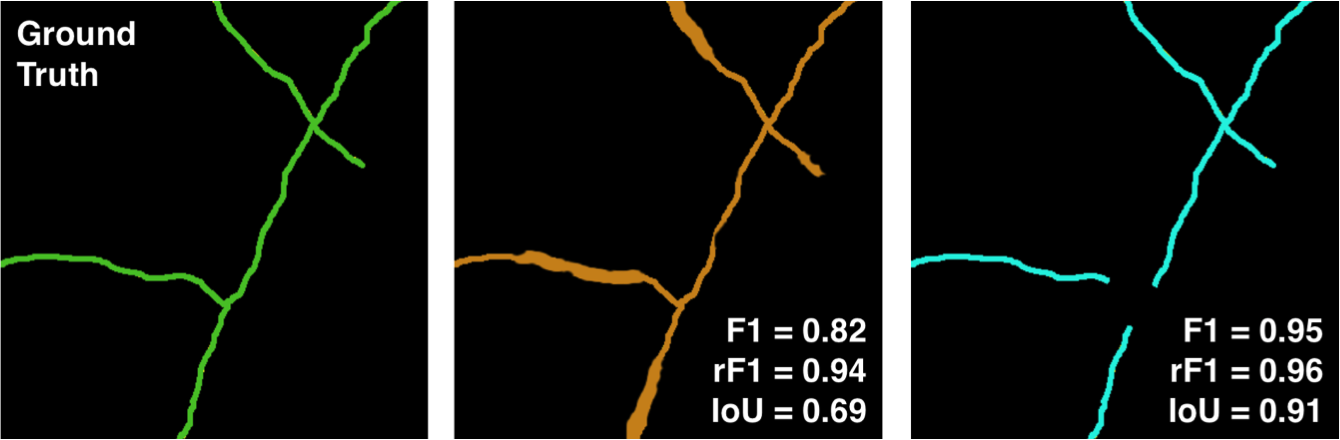}
  \caption{Legacy metrics often incentivize poor predictions.  Left: Ground truth road network in green.  Middle: proposal road mask in orange; the road widths are poorly reproduced, though there are no breaks or extraneous connections in the network, yielding scores of F1 = 0.82, relaxed F1 (rF1) = 0.94 for radius = 3), and IoU = 0.69.  Right: second proposal road mask in blue; road widths are correct, though a gap exists in the roads (often due to overhanging trees).  The right plot yields higher scores (F1 = 0.95, rF1 = 0.96 (radius = 3), and IoU = 0.91).  Therefore, legacy metrics prefer the right prediction to the middle prediction, even though the rightmost prediction would not be useful for routing purposes.}
  \label{fig:pix_metrics}
\end{figure}

The three metrics listed above all fail to adequately incentivize the creation of connected road networks. For example, for the IoU and F1 metrics a slight error in the road width is heavily penalized, though a brief break in an inferred road (from a shadow or overhanging tree) is lightly penalized, as illustrated in Figure \ref{fig:pix_metrics}.  

\subsection{APLS Metric}

Our focus on road networks naturally leads us to consider graph theory as a means to evaluate proposals. Most similarity measures in graph theory focus solely on the logical topology (connections between nodes) of the graph, whereas we care about both logical topology as well as the physical topology of the roads.  Nevertheless, metrics have been proposed specifically for road graph similarity matching.  \cite{topo_metric} proposed a metric for computing road graph similarity via comparing the nodes that can be reached within a small local vicinity of a number of seed nodes, categorizing proposal nodes as true positives, false positives, or false negatives depending on whether they fall within a buffer region (referred t to as the ``hole size''). By design, this metric evaluates local subgraphs in a small subregion ($\sim 300$ meters in extent), and relies upon physical geometry.  This metric is very sensitive to the locations of seed nodes, and connections between greatly disparate points ($>300$ meters apart) are not measured. 
\cite{5perc_metric} proposed computing the percentage of paths where the difference in distance between the ground truth and proposal path lengths was less than $5\%$; this metric gives some indication of length similarity, though provides little insight if differences are slightly greater than $5\%$, and does not penalize spurious unconnected proposals.

Since we are primarily interested in routing, (and to overcome some of the limitations of the metrics proposed in  \cite{topo_metric},  \cite{5perc_metric}) we propose a graph theoretic metric based upon Dijkstra's shortest path algorithm \cite{Dijkstra1959}. In essence, we symmetrically sum the differences in optimal paths between ground truth and proposal graphs. 

\begin{figure}
  \centering
     \includegraphics[width=0.7\linewidth]{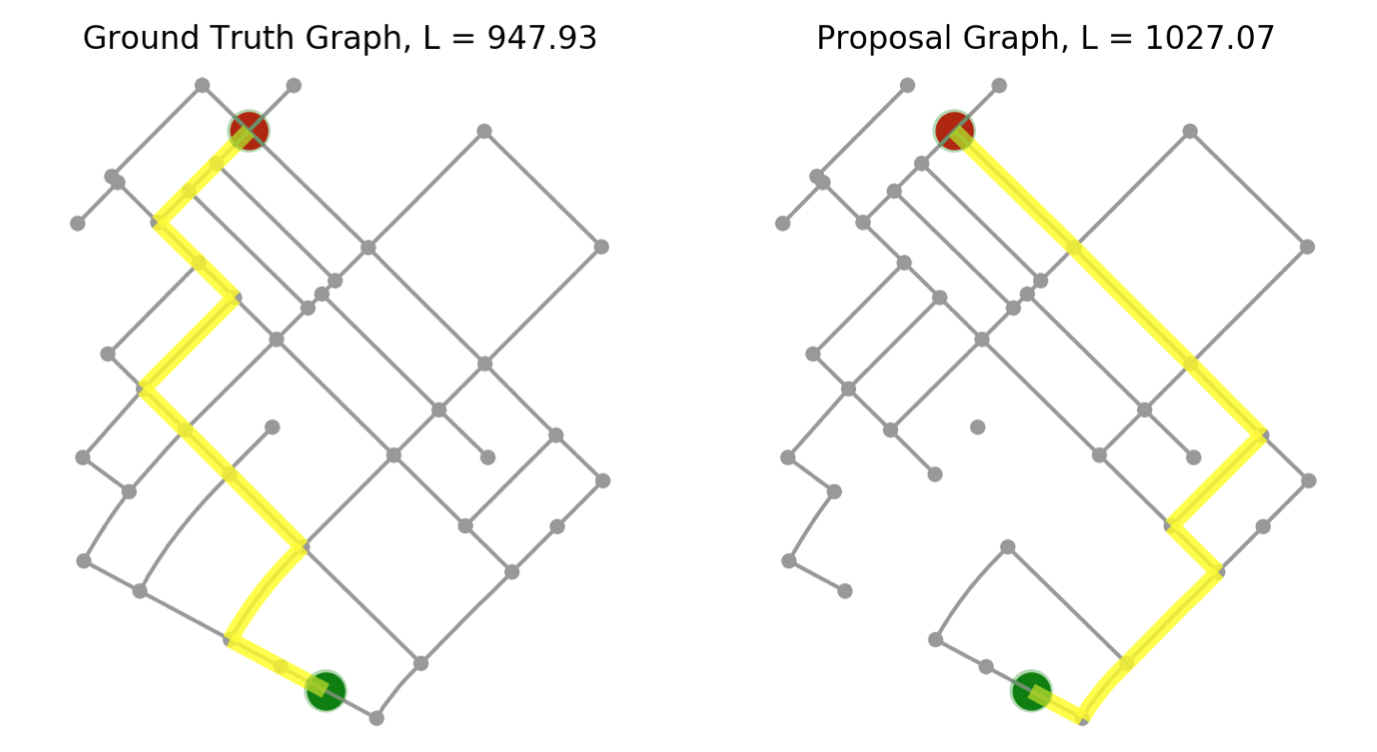}
  \caption{Demonstration of path length difference between sample ground truth and proposal graphs. Left: Shortest path between source (green) and target (red) node in the ground truth graph is shown in yellow, with a path length of $\approx948$ meters. Right: Shortest path between source and target node in the proposal graph with 30 edges removed, with a path length of $\approx1027$ meters; this difference in length forms the basis for our graph similarity metric. Plotting is accomplished via the osmnx python package \cite{osmnx}.
  }
  \label{fig:apls_fig0}
\end{figure}

To measure the difference between ground truth and proposal graphs,
 the APLS metric sums the differences in optimal path lengths between nodes in the ground truth graph G and the proposal graph G'. In effect, this metric repeats the path difference calculation shown in Figure \ref{fig:apls_fig0} for all paths in the graph. Missing paths in the graph are assigned the maximum proportional difference of 1.0. The APLS metric scales from 0 (poor) to 1 (perfect),

\begin{equation}
\label{eqn:apls}
C = 1 - \frac{1}{N} \sum min \left\{ 1, \frac{ \mid L(a,b) - L(a',b') \mid }{ L(a,b) }  \right\}
\end{equation}
where $N = $ number of unique paths, while $L(a,b) =$ length of path$(a,b)$.  The sum is taken over all possible source ($a$) and target ($b$) nodes in the ground truth graph.  The node $a'$ denotes the node in the proposal graph closest to the location of ground truth node $a$.  If path$(a',b')$ does not exist, the maximum contribution of 1.0 is used, thereby ensuring that missing routes are highly penalized. 


Inherent to this metric is the notion of betweenness centrality, or the number of times a node appears along the shortest paths between other nodes. Missing nodes of high centrality will be penalized much more heavily by the APLS metric than missing nodes of low centrality. This feature is intentional, as heavily trafficked intersections are much more important than cul-de-sacs for routing purposes. Any proposed graph G' with missing edges (e.g.~if an overhanging tree is inferred to sever a road) will be heavily penalized by the APLS metric, so ensuring that roads are properly connected is crucial for a high score. 

For small graphs the greedy approach of computing all possible paths in the network is entirely feasible 
(computing all possible paths for the 400m image chips of SpaceNet Challenge 3 takes less than 1 millisecond). 
For larger graphs, one must decide which nodes and paths are of paramount interest, lest the computational load become burdensome. 

While we developed the APLS metric as a means to score computer vision competitions, unlike pixel-based metrics the APLS metric is agnostic to data type.  The APLS metric applies equally well to road network proposals derived from optical imagery, radar, GPS tracks, LIDAR, and/or hand-crafted labels.  Unfortunately, this metric is not differentiable.

\subsubsection{Graph Augmentation}

We select control nodes in the graph to denote points and routes of interest.
In practice, control nodes are comprised of intersections, endpoints, and midpoints along edges.  In order to capture the physical topology of roads, we inject midpoint nodes every 50 meters.  

\subsubsection{Node Snapping}\label{sec:node_snap}

Equation \ref{eqn:apls} is easy to apply if the proposal graph G' has nodes coincident with the ground truth graph G. In practice, however, proposals will not align perfectly with ground truth graphs. Subsequently, we ``snap'' the control points of the ground truth graph onto the proposal graph. Proposal nodes must be within a given buffer distance of ground truth nodes.  This buffer is 4 meters by default, which is equivalent to the labeling accuracy requirement for the road centerlines.  Proposal nodes outside the buffer will not be snapped, and therefore highly penalized.  This process creates a new proposal graph with nodes as coincident as possible to the nodes of the ground truth graph.  Once the node snapping process is complete, routes between nodes in the ground truth graph (path$(a,b)$) can be directly compared to routes in the proposal graph (path$(a',b')$) using Equation \ref{eqn:apls}.
See Appendix C for more details.



\subsubsection{Symmetric Comparisons}

Once the ground truth to proposal node snapping procedure of Section \ref{sec:node_snap} is completed, optimal path lengths can be compared between proposal and ground truth nodes. In order to penalize spurious road proposals, one must also perform the inverse operation by snapping proposal control nodes onto the ground truth graph. This is illustrated in 
Appendix C.  


\subsubsection{Total APLS Metric}
APLS part 1 is computed by snapping ground truth control nodes onto the proposal graph and calculating Equation \ref{eqn:apls}.  APLS part 2 is computed via the reverse: snapping proposal nodes onto the ground truth graph and calculating Equation \ref{eqn:apls}. The total ALS metric is the harmonic mean of APLS part 1 and APLS part 2, and scales from 0 (poor) to 1 (perfect).

\section{SpaceNet Competition Results}

In this section we cover highlights of the winning submissions, the full results can be found on the SpaceNet website \cite{sn_web}.
All competitions were hosted on TopCoder.

\subsection{SpaceNet Challenge 1 Results - Rio de Janeiro Building Footprints}

For the first challenge, we baselined performance with a modified version of YOLO \cite{yolo9000} we call YOLT \cite{yolt}; this method achieved an F1 score of 0.21. This baseline provides a very different approach (and useful comparison) to the winning 3-class random forest classification + `polygonization' algorithm of wleite  \cite{wleite} (F1 = 0.26). One highlight of the insights provided by this initial challenge is the degree to which all algorithms struggled to distinguish individual row houses in the densely populated central regions of Rio de Janeiro.

\subsection{SpaceNet Challenge 2 Results - Las Vegas, Paris, Shanghai, Khartoum Building Footprints}

The second buildings challenge yielded far better results than the first.  This is likely due to a combination of three factors: increased resolution (30 cm vs 50 cm for SpaceNet Round 1), improved building labels (due to greater experience by the labeling team, higher resolution images, and lower density in the SpaceNet 2 cities), and/or improved competitor algorithms.  
We baselined performance with both YOLT (F1 = 0.60) and a modified version of MNC \cite{mnc} (F1 = 0.57).  
The winning algorithm submitted  by $XD\_XD$ \cite{xdxd} used an ensemble of three deep learning segmentation masks with a U-Net \cite{unet} architecture.  This ensemble, combined with thresholding and filtering yielded building polygon predictions and an F1 score of 0.69.
Las Vegas proved the easiest city to classify (partly due to the many well separated residential buildings with low variance in size), with Khartoum proving the most challenging (partly due to the high variance in building size and low contract between building and background).  See Table \ref{tab:sn2_res} for detailed results of the top three competitors, and Appendix D for images.

\begin{table}
  \caption{SpaceNet Challenge 2 Results (F1 Metric)}
  \label{tab:sn2_res}
  \centering
   \begin{tabular}{lcccccc}
    \toprule
    Rank & Competitor & Las Vegas & Paris & Shanghai & Khartoum & Total Score \\
    \hline
    1 & $XD\_XD$ & \bf{0.89} & \bf{0.75} & \bf{0.60} & \bf{0.54} & \bf{0.69} \\
    2 & wleite & 0.83 & 0.68 & 0.58 & 0.48 & 0.64 \\
    3 & nofto & 0.79 & 0.58 & 0.52 & 0.42 & 0.58 \\  
    \bottomrule
  \end{tabular}
\end{table}



\subsection{SpaceNet Challenge 3 Results - Las Vegas, Paris, Shanghai, Khartoum Road Networks}

\begin{figure}
  \centering
     \includegraphics[width=0.9\linewidth]{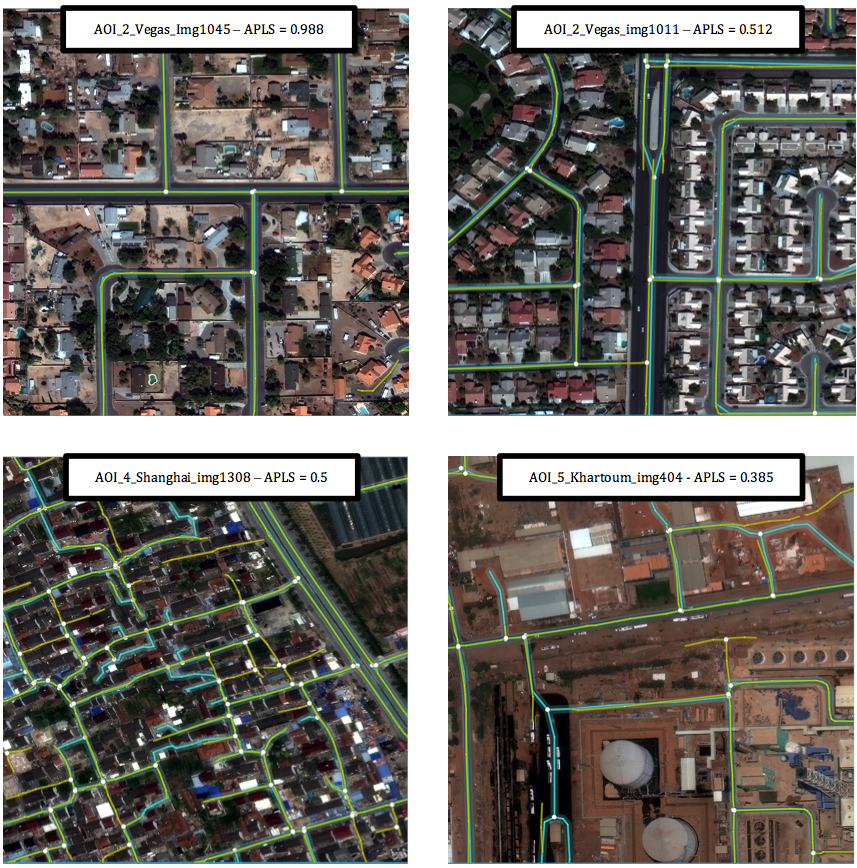}
  \caption{Results from the winning implementation for the roads challenge.  Top Left:  A simple road network in a 400m $\times$ 400m chip from the test set in Las Vegas; the blue line is the ground truth, the yellow line the proposal network, and the APLS score is 0.99. Top Right: A complex road network in Las Vegas; in the center of the graph network there is a disconnect where the divider is located.  Bottom Left: A complex network in Shanghai; there are several missed streets in the center of the graph. Bottom Right: A low scoring road network in Khartoum; the proposal network misses several dirt roads, but performs well on the more established paved road network.}
  \label{fig:sn3_res}
\end{figure}

Given the novelty of the metric used and the demand for geospatial vector submissions, entries to the first roads challenge were surprisingly good.
The top performers all took an approach similar to the baseline algorithm (APLS = 0.49) outlined in Appendix E (i.e. U-Net + skeletonization + sknw \cite{sknw}), but with greater attention paid to network architecture and post-processing. 
The winning algorithm was submitted by competitor albu \cite{albu},  which used an ensemble of deep learning segmentation encoders/decoders to train a global road model.  Results from the winning algorithm are shown in Figure \ref{fig:sn3_res}, and detailed results of the top five competitors is shown in Table \ref{tab:sn3_res} (we show the top five competitors and three significant figures due to the closeness of scores).

\begin{table}
  \caption{SpaceNet Challenge 3 Results (APLS Metric)}
  \label{tab:sn3_res}
  \centering
   \begin{tabular}{lcccccc}
    \toprule
    Rank & Competitor & Las Vegas & Paris & Shanghai & Khartoum & Total Score \\
    \hline
    1 & albu 	  & 0.798 & 0.604 & 0.654 & \bf{0.609} & \bf{0.6663} \\
    2 & cannab 	  & 0.780 & \bf{0.645} & 0.640 & 0.600 & 0.6661 \\
    3 & pfr 		  & \bf{0.801} & 0.601 & \bf{0.665} & 0.598 & 0.666 \\  
    4 & selim\_sef & 0.788 & 0.599 & 0.647 & 0.592 & 0.657 \\  
    5 & fabastani  & 0.771 & 0.547 & 0.633 & 0.563 & 0.628 \\  
    \bottomrule
  \end{tabular}
\end{table}

\section{Conclusions}\label{conclusions}

Foundational mapping has myriad applications, from commercial (e.g.~autonomous vehicles), to humanitarian (e.g.~disaster response), to civil (e.g.~traffic congestion mitigation).  
Satellite imagery may be able to aid foundational mapping, particularly in cases of natural disasters or other dynamic events where the revisit rate of satellites may be able to provide updates far faster than terrestrial or airborne methods, and cover a large geographic area.

The SpaceNet dataset provides a large corpus of high resolution multi-band imagery, with attendant validated building footprint and road network labels.  The SpaceNet building footprint extraction challenges have yielded algorithms with F1 scores of $0.69$ on multiple cities, and vast improvement in performance from Challenge 1 to Challenge 2.    

In this paper, we demonstrated that the standard technique for scoring road detection algorithms (pixel-based F1 score) is suboptimal for routing purposes. Accordingly, we developed the graph-theoretic APLS metric based on shortest paths between nodes in ground truth and proposal networks. This metric rewards correct road centerline and intersection determination to a far greater extent than pixel-based metrics.  The SpaceNet competition yielded multiple high quality submissions in graph structure format, and a winning score of 0.66.  These outputs can be used directly for routing, and provide a step towards addressing the many problems that improved automated routing impacts.  

We will continue adding imagery and labels to the SpaceNet data corpus, as well as hosting challenges to encourage the development of algorithms tailored to the unique aspects of satellite imagery.  
The next (fourth) challenge will analyze the ability of algorithms to correctly localize building footprints in off-nadir (i.e. oblique look angle) imagery; this capability is sorely needed in many disaster response scenarios, where the first available image of an area may be at a very high off-nadir angle. The fifth SpaceNet competition will revisit road network extraction, this time utilizing metadata to explore road network travel times. Subsequent challenges will use metadata collected in the ongoing labeling campaigns to methodically increase problem complexity and difficulty.
Our aspiration is that the large amount of labeled SpaceNet data will stir the imagination of an increasing cadre of computer vision experts, and lead to improvements in remote sensing analytics.




\clearpage

\bibliographystyle{splncs}
\bibliography{bib}

\begin{thebibliography}{10}

\bibitem{sn_s3}
AWS:
\newblock Spacenet on aws.
\newblock \url{https://registry.opendata.aws/spacenet/} (2017)

\bibitem{zhangpop}
Zhang, F., Du, B., Zhang, L.:
\newblock A multi-task convolutional neural network for mega-city analysis
  using very high resolution satellite imagery and geospatial data.
\newblock CoRR \textbf{abs/1702.07985} (2017)

\bibitem{popnature}
Tatem, A.J.:
\newblock Worldpop, open data for spatial demography.
\newblock Scientific Data \textbf{4} (01 2017)  170004 EP --

\bibitem{popplos1}
Stevens, F.R., Gaughan, A.E., Linard, C., Tatem, A.J.:
\newblock Disaggregating census data for population mapping using random
  forests with remotely-sensed and ancillary data.
\newblock PLOS ONE \textbf{10}(2) (02 2015)  1--22

\bibitem{hotosm}
HOTOSM:
\newblock Hot tasking manager.
\newblock \url{https://tasks.hotosm.org/} (2018)

\bibitem{imagenet}
Russakovsky, O., Deng, J., Su, H., Krause, J., Satheesh, S., Ma, S., Huang, Z.,
  Karpathy, A., Khosla, A., Bernstein, M., Berg, A.C., Fei-Fei, L.:
\newblock {ImageNet Large Scale Visual Recognition Challenge}.
\newblock International Journal of Computer Vision (IJCV) \textbf{115}(3)
  (2015)  211--252

\bibitem{isprs_sem}
ISPRS:
\newblock 2d semantic labeling contest.
\newblock
  \url{http://www2.isprs.org/commissions/comm3/wg4/semantic-labeling.html}
  (2018)

\bibitem{torontocity}
Wang, S., Bai, M., M{\'{a}}ttyus, G., Chu, H., Luo, W., Yang, B., Liang, J.,
  Cheverie, J., Fidler, S., Urtasun, R.:
\newblock Torontocity: Seeing the world with a million eyes.
\newblock CoRR \textbf{abs/1612.00423} (2016)

\bibitem{MnihThesis}
Mnih, V.:
\newblock Machine Learning for Aerial Image Labeling.
\newblock PhD thesis, University of Toronto (2013)

\bibitem{cowc}
Mundhenk, T.N., Konjevod, G., Sakla, W.A., Boakye, K.:
\newblock A large contextual dataset for classification, detection and counting
  of cars with deep learning.
\newblock CoRR \textbf{abs/1609.04453} (2016)

\bibitem{xview_data}
Lam, D., Kuzma, R., McGee, K., Dooley, S., Laielli, M., Klaric, M., Bulatov,
  Y., McCord, B.:
\newblock xview: Objects in context in overhead imagery.
\newblock CoRR \textbf{abs/1802.07856} (2018)

\bibitem{dstl}
Kaggle:
\newblock Dstl satellite imagery feature detection.
\newblock
  \url{https://www.kaggle.com/c/dstl-satellite-imagery-feature-detection}
  (2016)

\bibitem{fmow}
Christie, G., Fendley, N., Wilson, J., Mukherjee, R.:
\newblock Functional map of the world.
\newblock CoRR \textbf{abs/1711.07846} (2017)

\bibitem{xview}
Lam, D., Kuzma, R., McGee, K., Dooley, S., Laielli, M., Klaric, M., Bulatov,
  Y., McCord, B.:
\newblock xview: Objects in context in overhead imagery.
\newblock CoRR \textbf{abs/1802.07856} (2018)

\bibitem{deepglobe}
DeepGlobe:
\newblock Deepglobe - cvpr 2018.
\newblock \url{http://deepglobe.org/challenge.html} (2018)

\bibitem{isprssemlab}
Gerke, M., Rottensteiner, F., D~Wegner, J., Sohn, G.:
\newblock Isprs semantic labeling contest (09 2014)

\bibitem{sn_utils}
CosmiQWorks:
\newblock Spacenet challenge utilites.
\newblock \url{https://github.com/SpaceNetChallenge/utilities/} (2018)

\bibitem{Urban3D2017}
Goldberg, H., Brown, M., Wang, S.:
\newblock A benchmark for building footprint classification using
  orthorectified rgb imagery and digital surface models from commercial
  satellites.
\newblock In: Proceedings of IEEE Applied Imagery Pattern Recognition Workshop
  2017. (2017)

\bibitem{apls}
CosmiQWorks:
\newblock Apls metric.
\newblock \url{https://github.com/CosmiQ/apls} (2017)

\bibitem{relaxedf1}
Mnih, V., Hinton, G.E.:
\newblock Learning to detect roads in high-resolution aerial images.
\newblock In: Proceedings of the 11th European Conference on Computer Vision:
  Part VI. ECCV'10, Berlin, Heidelberg, Springer-Verlag (2010)  210--223

\bibitem{topo_metric}
Biagioni, J., Eriksson, J.:
\newblock Inferring road maps from global positioning system traces: Survey and
  comparative evaluation.
\newblock Transportation Research Record \textbf{2291}(1) (2012)  61--71

\bibitem{5perc_metric}
Wegner, J.D., Montoya-Zegarra, J.A., Schindler, K.:
\newblock A higher-order crf model for road network extraction.
\newblock 2013 IEEE Conference on Computer Vision and Pattern Recognition
  (2013)  1698--1705

\bibitem{Dijkstra1959}
Dijkstra, E.W.:
\newblock A note on two problems in connexion with graphs.
\newblock Numerische Mathematik \textbf{1}(1) (Dec 1959)  269--271

\bibitem{osmnx}
Boeing, G.:
\newblock Osmnx: New methods for acquiring, constructing, analyzing, and
  visualizing complex street networks.
\newblock Computers, Environment and Urban Systems \textbf{65} (2017)  126 --
  139

\bibitem{sn_web}
CosmiQWorks:
\newblock Spacenet on aws.
\newblock \url{https://spacenetchallenge.github.io/} (2018)

\bibitem{yolo9000}
Redmon, J., Farhadi, A.:
\newblock {YOLO9000:} better, faster, stronger.
\newblock CoRR \textbf{abs/1612.08242} (2016)

\bibitem{yolt}
{{Van Etten}}, A.:
\newblock You only look twice: Rapid multi-scale object detection in satellite
  imagery.
\newblock Submitted to KDD (2018)

\bibitem{wleite}
wleite:
\newblock Spacenet round 1 winner: wleite's implementation.
\newblock
  \url{https://github.com/SpaceNetChallenge/BuildingDetectors/tree/master/wleite}
  (2017)

\bibitem{mnc}
Dai, J., He, K., Sun, J.:
\newblock Instance-aware semantic segmentation via multi-task network cascades.
\newblock CoRR \textbf{abs/1512.04412} (2015)

\bibitem{xdxd}
xd\_xd:
\newblock Spacenet round 2 winner: Xd\_xd's implementation.
\newblock
  \url{https://github.com/SpaceNetChallenge/BuildingDetectors_Round2/tree/master/1-XD
  \_XD} (2017)

\bibitem{unet}
Ronneberger, O., Fischer, P., Brox, T.:
\newblock U-net: Convolutional networks for biomedical image segmentation.
\newblock CoRR \textbf{abs/1505.04597} (2015)

\bibitem{sknw}
yxdragon:
\newblock Skeleton network.
\newblock \url{https://github.com/yxdragon/sknw} (2018)

\bibitem{albu}
albu:
\newblock Spacenet round 3 winner: albu's implementation.
\newblock
  \url{https://github.com/SpaceNetChallenge/RoadDetector/tree/master/albu-solution}
  (2018)

\bibitem{basiss}
CosmiQWorks:
\newblock Broad area satellite imagery semantic segmentation.
\newblock \url{https://github.com/CosmiQ/basiss} (2018)

\bibitem{pspnet}
Zhao, H., Shi, J., Qi, X., Wang, X., Jia, J.:
\newblock Pyramid scene parsing network.
\newblock CoRR \textbf{abs/1612.01105} (2016)

\bibitem{skimage}
scikit image:
\newblock Skeletonize.
\newblock
  \url{https://scikit-image.org/docs/dev/auto_examples/edges/plot_skeleton.html}
  (2018)

\end{thebibliography}

\setcounter{secnumdepth}{0}

\newpage

\section{Appendix A.  SpaceNet Building Footprint Labeling Guide}

\begin{enumerate}
	\item Buildings will primarily be within urban/suburban areas.
	\item Rooflines will first be extracted to best represent the shape of the footprint and for any buildings where the base of the building is partially visible due to off-nadir angle the polygon will be shifted to best fit the base corners.
	\item Buildings will have squared corners (assuming they are true to the shape of the building).
	\item Extraction of building corners will be within 5 pixels of the location on the imagery.
	\item Building footprints will be closed polygons and will not overlap.
	\item No attributes will be populated.
	\item Adjoining buildings with different heights shall be captured at one visible roof for the whole block and shifted to that visible roofs baseline.
\end{enumerate}

\newpage

\section{Appendix B.  The SpaceNet Roads Dataset Labeling Guidelines}

\subsection{The SpaceNet Roads Dataset labeling guidelines}

\begin{enumerate}
	\item Road vectors must be drawn as a center line within 2m (7 pixels) of observed road
	\begin{enumerate}
		\item The centerline of a road is defined as the centerline of the roadway. If a road has an even number of lanes, the centerline shall be drawn on the line separating lanes.  If the road has an odd number of lanes then the centerline should be drawn down the center of the middle lane.  
		\item Divided highways should have two centerlines, a centerline for each direction of traffic.  See below for the definition of a divided highway. 
	\end{enumerate}
	\item Road vectors must be represented as a connected network to support routing.  Roads that intersect each other should share points as an intersection like instructed through OSM.  Roads that cross each other that are not connected such as while using an overpass should not share a point of connection.
	\item Roads must not bisect building footprints.
	\item Sections of a road that are a bridge or overpass must be labeled as a bridge via a Boolean flag.  
	\item Divided highways must be represented as two lines with traffic direction indicated when possible.
	\item Surface type must be classified as:  paved, unpaved, or unknown. 
	\item Road will be identified by type: (Motorway, Primary, Secondary, Tertiary, Residential, Unclassified, Cart Track)
	\item Number of lanes will be listed as number of lanes for each centerline as defined in rule 1.  If road has two lanes in each direction, the number of lanes shall equal 4.  If a road has 3 lanes in one direction and 2 directions in another the total number of lanes shall be 5.  
\end{enumerate}

\subsection{Definition of Divided Highway}

A divided highway is a road that has a median or barrier that physically prevents turns across traffic.  
A median can be:
\begin{itemize}
	\item Concrete
	\item Asphalt
	\item Green Space
	\item Dirt/unpaved road
\end{itemize}

A median is not:
\begin{itemize}
	\item Yellow hatched lines on pavement.  
\end{itemize}

\subsection{Road Type Guidelines}

All road types were defined using the Open Street Maps taxonomy for key=highway. The below descriptions were taken from the Open Street Maps tagging guidelines for highway and the East Africa Tagging Guidelines.  
\begin{enumerate}
	\item  motorway - A restricted access major divided highway, normally with 2 or more running lanes plus emergency hard shoulder. Access onto a motorway comes exclusively through ramps (controlled access).  Equivalent to the Freeway, Autobahn, etc.
	\item primary - National roads connect the most important cities/towns in a country. In most countries, these roads will usually be tarmacked and show center markings. (In South Sudan, however, primary roads might also be unpaved.)
	\item secondary - Secondary roads are the second most important roads in a country's transport system. They typically link medium-sized places. They may be paved but in in some countries they are not.
	\item tertiary - Tertiary roads are busy through roads that link smaller towns and larger villages. More often than not, these roads will be unpaved. However, this tag should be used only on roads wide enough to allow two cars to pass safely.
	\item residential - Roads which serve as an access to housing, without function of connecting settlements. Often lined with housing.
	\item unclassified - The least important through roads in a country's system - i.e. minor roads of a lower classification than tertiary, but which serve a purpose other than access to properties. Often link villages and hamlets. (The word ``unclassified'' is a historical artifact of the UK road system and does not mean that the classification is unknown; you can use highway=road for that.)
	\item Cart track - This is a dirt path that shows vehicle traffic that is less defined than a residential   
\end{enumerate}
	

\subsection{Road Label GeoJSON Schema}
Attributes:
\begin{enumerate}
	\item ``geometry'': Linestring
	\item ``road\_id'': int; Identifier Index
	\item ``road\_type'': int
	\begin{enumerate}
		\item Motorway
		\item Primary
		\item Secondary
		\item Tertiary
		\item Residential
		\item Unclassified
		\item Cart track
	\end{enumerate}
	\item ``paved'': int
	\begin{enumerate}
		\item Paved
		\item Unpaved
		\item Unknown
	\end{enumerate}
	\item ``bridge\_type'': int
	\begin{enumerate}
		\item Bridge
		\item Not a bridge
		\item Unknown
	\end{enumerate}
	\item ``lane\_number'': int
	\begin{enumerate}
		\item one lane
		\item two lane
		\item three lanes
		\item etc.
	\end{enumerate}
\end{enumerate}

\newpage

\section{Appendix C.   APLS Metric}

\begin{figure}[h]
  \centering
     \includegraphics[width=0.99\linewidth]{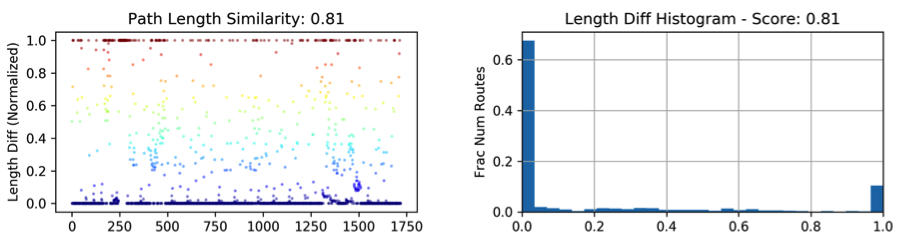}
  \caption{Left: Path length differences for all 1720 possible routes possible in the ground truth graph of Figure \ref{fig:apls_fig0}. Right: Path length difference histogram. Most routes are the same between the proposal and ground truth graphs, resulting in the many blue dots on the left plot and the large spike at path length 0.0 on the right plot. There are a few missing routes due to the disconnected node in the center of the proposal graph, and this results in the red dots on the left plot and the small spike at path length 1.0 on the right plot. Applying Equation 5 to this data yields APLS = 0.81.}
  \label{fig:apls_fig3}
\end{figure}

Path length differences are summed for each possible route in the graphs, which we illustrate in Figures \ref{fig:apls_fig0}, \ref{fig:apls_fig3}.

\begin{figure}[h]
  \centering
     \includegraphics[width=0.99\linewidth]{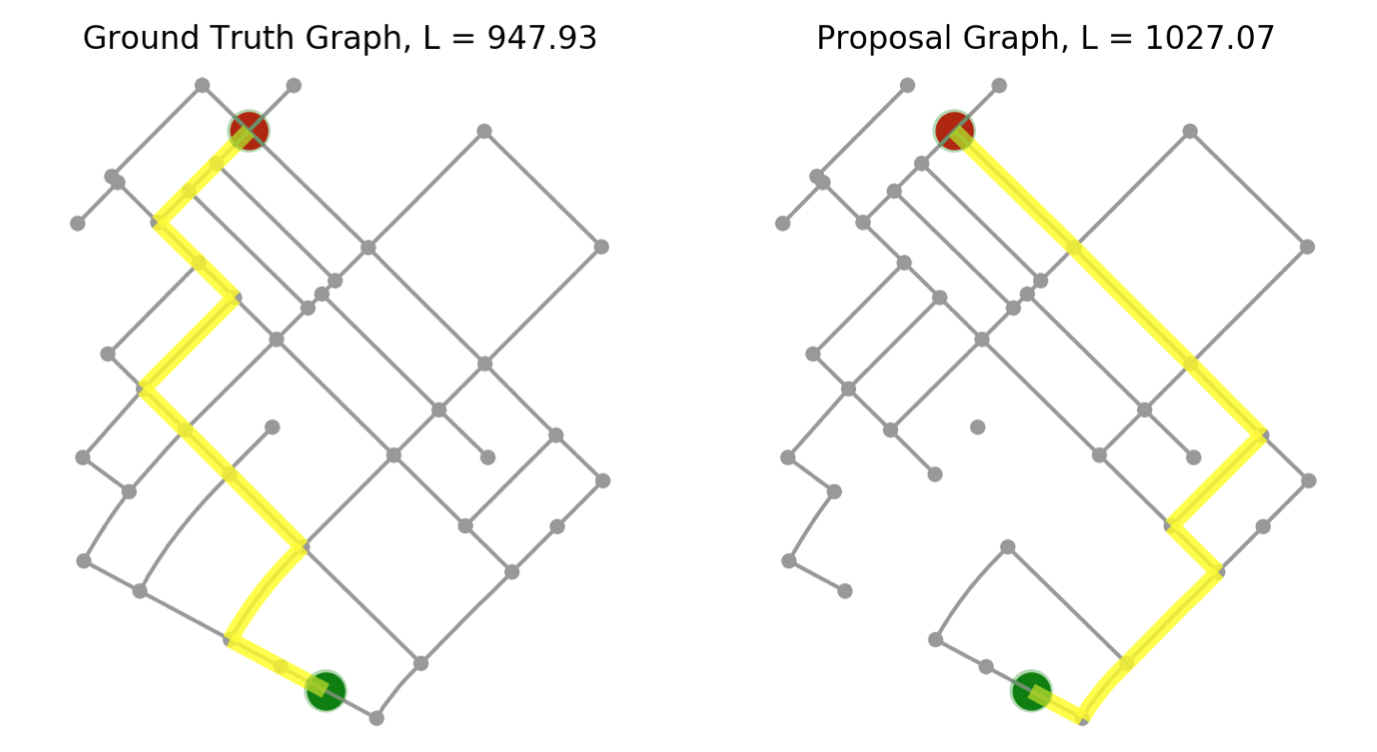}
  \caption{Demonstration of path length difference between sample ground truth and proposal graphs. Left: Shortest path between source (green) and target (red) node in the ground truth graph is shown in yellow, with a path length of ~948 meters. Right: Shortest path between source and target node in the proposal graph with 30 edges removed, with a path length of ~1027 meters; this difference in length forms the basis for our graph similarity metric. Plotting is accomplished via the osmnx python package.}
  \label{fig:apls_fig0}
\end{figure}

\subsection{APLS Graph Augmentation}
\begin{figure}[h]
  \centering
     \includegraphics[width=0.99\linewidth]{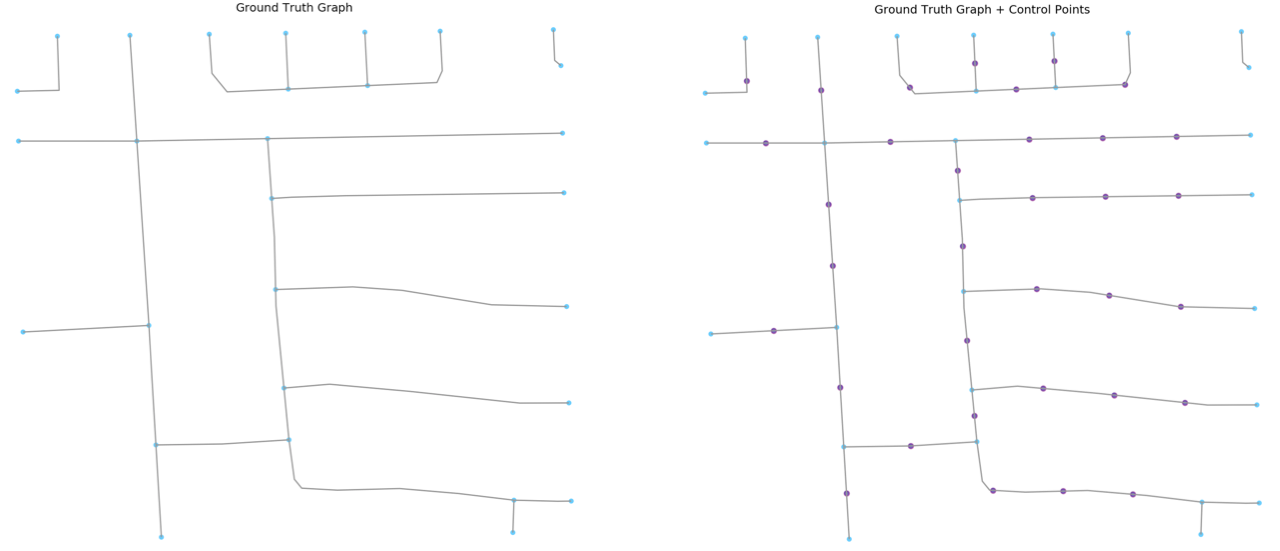}
  \caption{Left: Ground truth graph.  Right: Ground truth graph with control nodes (purple) injected along routes.  The graph at right is the one used to compute the APLS metric. }
  \label{fig:apls_fig5}
\end{figure}

A couple of complications arise when comparing road networks, both relating to the selection of which routes to compare. Accordingly, we select ``control nodes'' in the graph between which routes are of interest.  In practice, control nodes are comprised of intersections, endpoints, and midpoints along edges.  In order to capture the physical topology of roads, we inject nodes every 50 meters.  Obviously, for very large graphs one would be far more parsimonious with the selection of control nodes, but for the 400m SpaceNet tiles we can afford to select a high density of control nodes.  This process is shown in Figure  \ref{fig:apls_fig5}.

\subsection{Node Snapping}

Equation 3 is easy to apply if the proposal graph G' has nodes coincident with the ground truth graph G. In practice, however, proposals will not align perfectly with ground truth graphs. Subsequently, we ``snap'' the control points of the ground truth graph onto the proposal graph. Proposal nodes must be within a given buffer distance of ground truth nodes.  This buffer is 4 meters by default, which is equivalent to the labeling accuracy requirement.  Proposal nodes outside the buffer will not be snapped, and therefore highly penalized.  This process creates a new proposal graph with nodes as coincident as possible to the nodes of the ground truth graph.  Once the node snapping process is complete, routes between nodes in the ground truth graph (path$(a,b)$) can be directly compared to routes in the proposal graph (path$(a',b')$ using Equation 3.  
See Appendix A for more details.

\begin{figure}[h]
  \centering
     \includegraphics[width=0.99\linewidth]{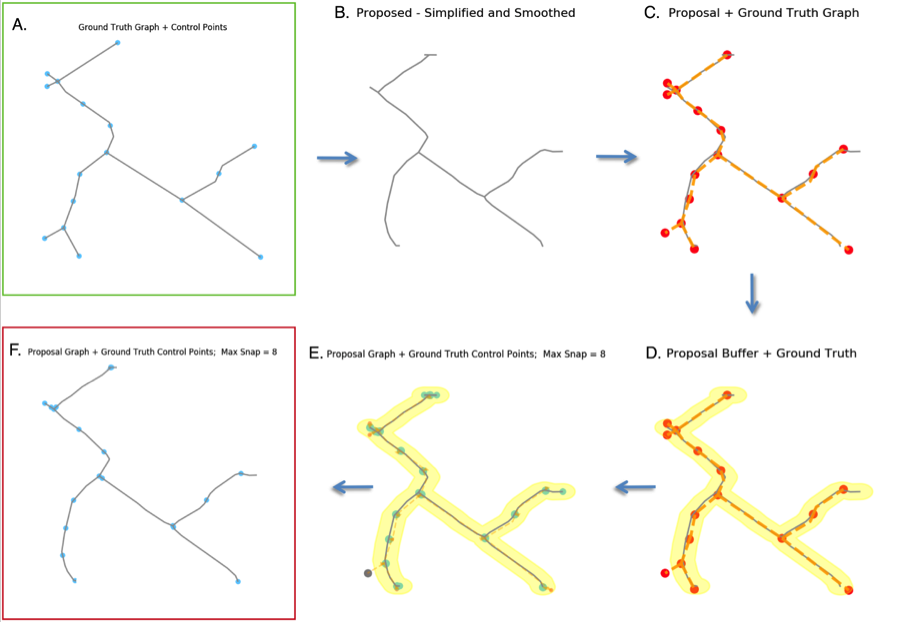}
  \caption{Node snapping procedure. A: Ground truth graph with control nodes. B: Proposal graph. C: Ground truth graph (orange) and ground truth control points (red) overlaid on proposal graph (grey). D: Same as graph C, but showing the buffer (yellow) around the proposal graph. E: Ground truth control nodes are injected into the proposal graph at the nearest edge, except for nodes outside the buffer (grey). F: Final proposal graph with nodes injected at the appropriate location to compare to graph A.}
  \label{fig:apls_fig1}
\end{figure}

Once the process of Figure \ref{fig:apls_fig1} is complete, routes between nodes of  the ground truth graph with control points (Figure \ref{fig:apls_fig1}A) and the augmented proposal graph (Figure \ref{fig:apls_fig1}F) can be directly compared.

\subsection{Symmetric Comparisons}

\begin{figure}[h]
  \centering
     \includegraphics[width=0.99\linewidth]{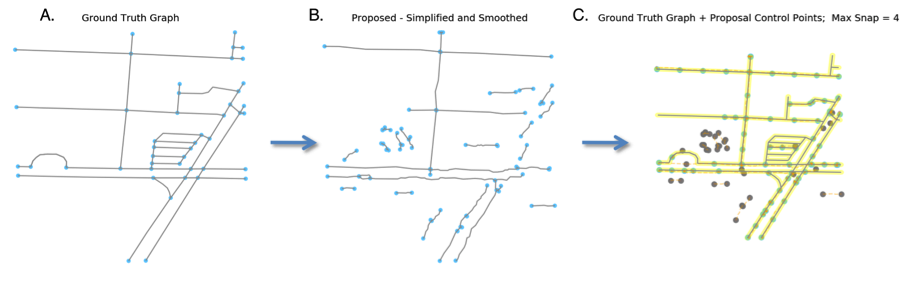}
  \caption{Illustration of the need to apply Equation 3 symmetrically (i.e.: ground truth to proposal, and proposal to ground truth). A: Ground truth graph. B: Proposal graph with many short, spurious connections well outside the buffer. C: Proposal nodes snapped onto the ground truth graph; snapping proposal control points onto the ground truth graph and then computing the metric penalizes the many extraneous predictions (grey nodes).}
  \label{fig:apls_fig2}
\end{figure}

Once the ground truth to proposal node snapping procedure of Figure \ref{fig:apls_fig1} is completed, optimal path lengths can be compared between proposal and ground truth nodes. In order to penalize spurious road proposals, one must also perform the inverse operation by snapping proposal control nodes onto the ground truth graph. This is illustrated in Figure \ref{fig:apls_fig2}.

\clearpage
\newpage

\section{Appendix D.  Building Detection Results}

Here we show a few snapshots of the winning algorithm for SpaceNet Challenge 2.

\begin{figure}[h]
  \centering
     \includegraphics[width=0.99\linewidth]{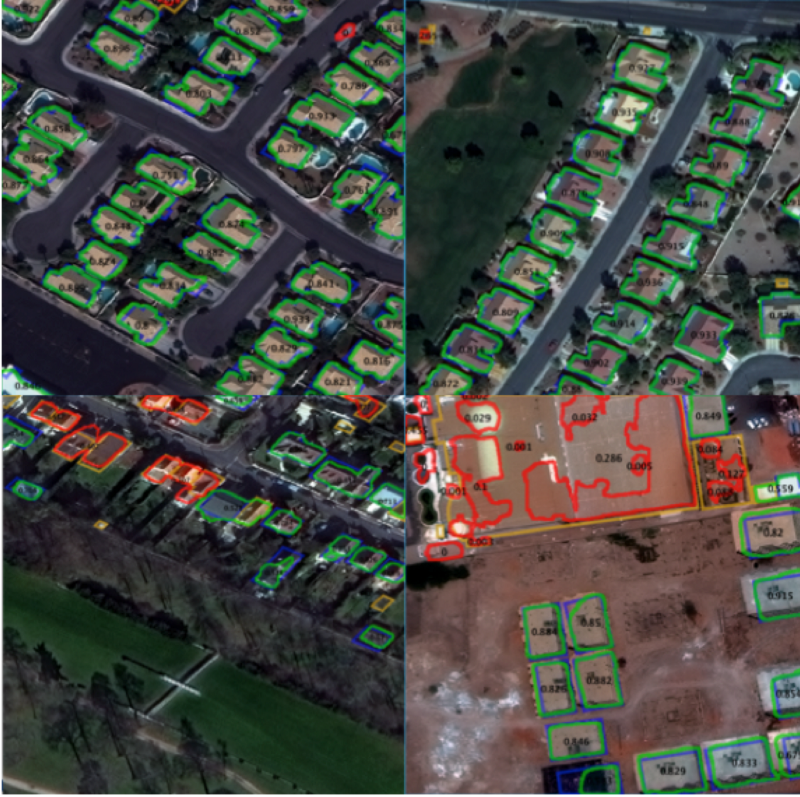}
  \caption{Results from the winning implementation of $XD\_XD$ for Buildings Round 2: From top left, Clockwise (Vegas, Vegas, Khartoum, Paris). The blue outline represents the ground truth, the green outlines are true positives, the red are false positives.}
  \label{fig:sn2_spacenet}
\end{figure}

\clearpage
\newpage

\section{Appendix E.  Road Baseline Algorithm}
The APLS metric is designed to reward correct node placement and connectivity, and so should prove a better metric than pixel-based F1 for automated route inference evaluation. In this section, we discuss the CosmiQ baseline procedures for exploring the application of APLS to satellite imagery \cite{basiss}.

\begin{figure}[h]
  \centering
     \includegraphics[width=0.99\linewidth]{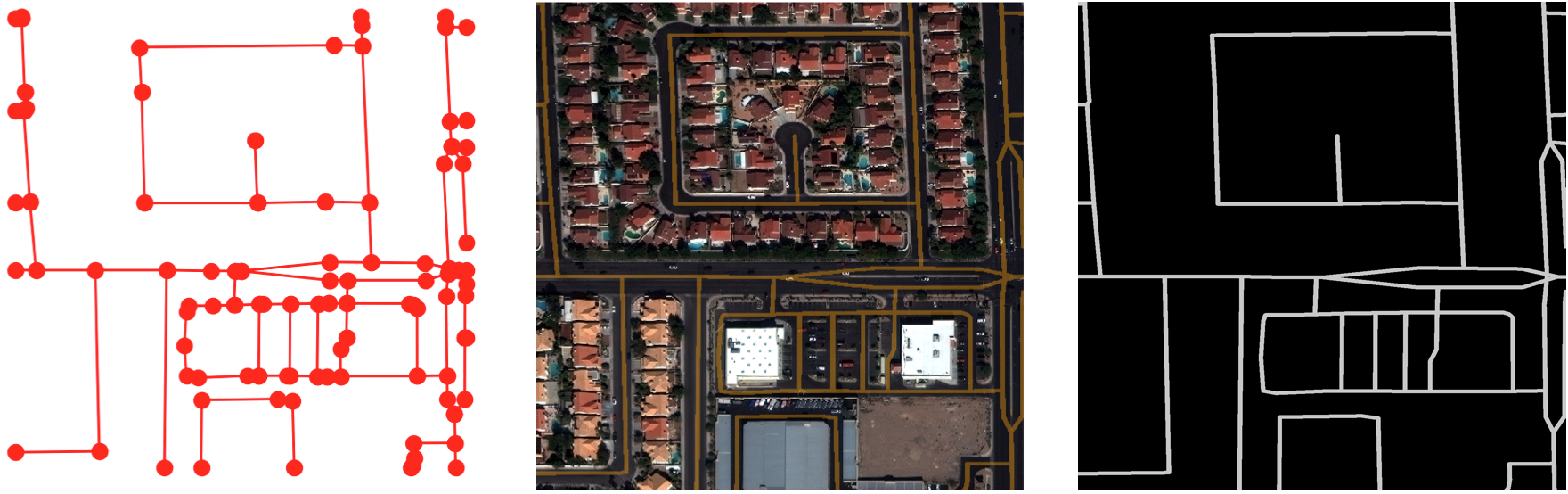}
  \caption{Left: SpaceNet GeoJSON label.  Middle: Image with road network overlaid in orange.  Right. Segmentation mask of road centerlines.}
    \label{fig:baseline_train}
\end{figure}

While any number of approaches are valid (and encouraged) to tackle the SpaceNet challenge, the most obvious workflow begins with inferring a segmentation mask using convolutional neural networks (CNNs).  We assume a road centerline halfwidth of 2m and create training masks using the raw imagery and SpaceNet geojson road labels (see Figure \ref{fig:baseline_train}).

\begin{figure}[h]
  \centering
     \includegraphics[width=0.99\linewidth]{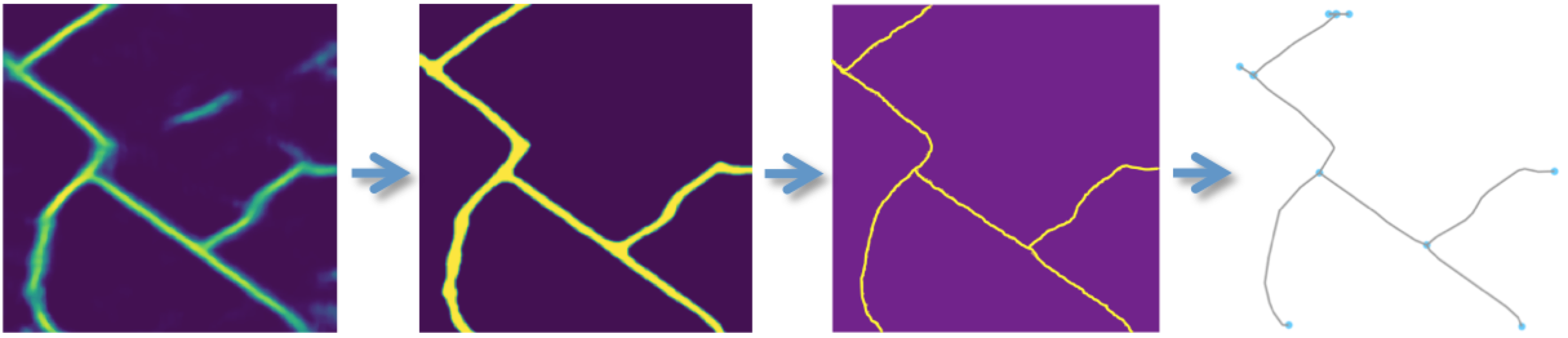}
  \caption{Baseline algorithm.  Left: Using road masks, we train CNN segmentation algorithms (such as PSPNet \cite{pspnet} and U-Net \cite{unet}) to infer road masks from SpaceNet imagery.  Left center: These outputs masks are then refined using standard techniques such as thresholding, opening, closing, and smoothing.  Right center: A skeleton is created from this refined mask (e.g.: sckit-image skeletonize \cite{skimage}.
  Right: This skeleton is subsequently rendered into a graph structure, such as with the sknw package \cite{sknw}.  
  Competitors to the roads challenge generally followed the same approach as this baseline algorithm, albeit with different network architectures and more refined post-processing steps.
}
    \label{fig:baseline}
\end{figure}

\begin{figure}[h]
  \centering
     \includegraphics[width=0.99\linewidth]{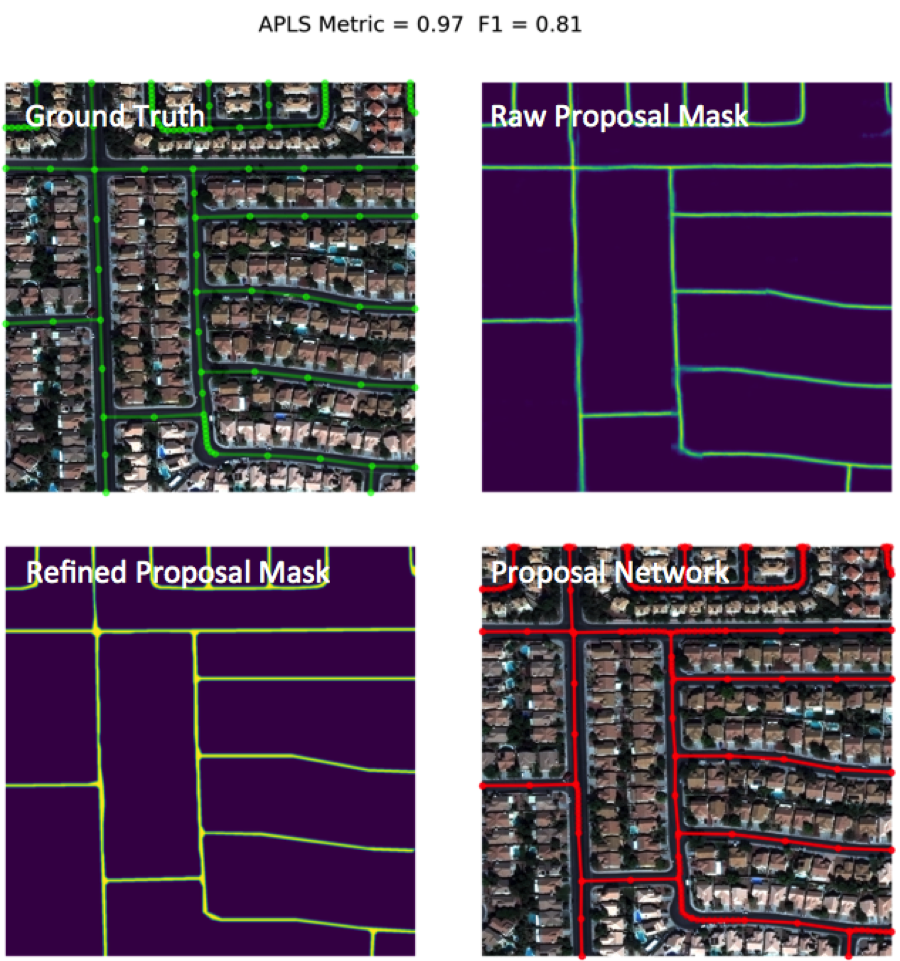}
  \caption{Sample output of baseline algorithm applied to SpaceNet test data.}
    \label{fig:baseline_ex}
\end{figure}

\end{document}